\documentclass[10pt,twocolumn,letterpaper]{article}
\pdfoutput=1 
\usepackage{iccv}
\usepackage{times}
\usepackage{epsfig}
\usepackage{graphicx}
\usepackage{amsmath}
\usepackage{amsfonts}
\usepackage{subfig}

\usepackage{xspace}
\usepackage{makecell}
\usepackage{colortbl}
\usepackage{multirow}
\usepackage{paralist}

\newcommand{\mypartitle}[2][2.4]{\vspace*{-#1 ex}~\\{\noindent {\bf #2}}}

\makeatletter
\DeclareRobustCommand\onedot{\futurelet\@let@token\@onedot}
\def\@onedot{\ifx\@let@token.\else.\null\fi\xspace}

\def\eg{{e.g}\onedot}

\def\etc{{etc}\onedot}

\def\soa{{state-of-the-art}\xspace}
\def\potr{{POTR}\xspace}
\makeatother

\definecolor{Gray}{gray}{0.9} 

\newcommand{\EmbDimension}[0]{\ensuremath{D}\xspace}
\newcommand{\TransParams}[0]{\ensuremath{\theta}\xspace}
\newcommand{\SelfAttnEmb}[0]{\ensuremath{\mathbf{z}}\xspace}
\newcommand{\InputPose}[0]{\ensuremath{\mathbf{x}}\xspace}
\newcommand{\PredictedPose}[0]{\ensuremath{\mathbf{y}}\xspace}
\newcommand{\QueryPose}[0]{\ensuremath{\mathbf{q}}\xspace}
\newcommand{\PoseDimension}[0]{\ensuremath{N}\xspace}
\newcommand{\TargetLength}[0]{\ensuremath{M}\xspace}
\newcommand{\InputLength}[0]{\ensuremath{T}\xspace}
\newcommand{\PoseEmbFn}[0]{\ensuremath{\phi}\xspace}
\newcommand{\PoseDecFn}[0]{\ensuremath{\psi}\xspace}
\newcommand{\NumLayers}[0]{\ensuremath{L}\xspace}
\newcommand{\NumNodes}[0]{\ensuremath{K}\xspace}
\newcommand{\AdjMatrix}[0]{\ensuremath{\mathbf{A}}\xspace}
\newcommand{\NodeFeats}[0]{\ensuremath{\mathbf{H}}\xspace}
\newcommand{\NodeWeights}[0]{\ensuremath{\mathbf{W}}\xspace}
\newcommand{\NumNodeFeats}[0]{\ensuremath{F}\xspace}
\newcommand{\NodeNumFeatsOut}[0]{\ensuremath{O}\xspace}

\usepackage[pagebackref=true,breaklinks=true,letterpaper=true,colorlinks,bookmarks=false]{hyperref}

\iccvfinalcopy 

\def\iccvPaperID{7} 

\ificcvfinal\fi

\def\idiapref{\textsuperscript{\textdagger}}
\def\epflref{\textsuperscript{\textasteriskcentered}}

\begin{document}

\title{Pose Transformers (POTR): Human Motion Prediction with\\Non-Autoregressive Transformers}

\author{Angel Mart\'inez-Gonz\'alez\epflref\idiapref,
Michael Villamizar\idiapref
and
Jean-Marc Odobez\epflref\idiapref\\
\epflref\'Ecole Polytechnique F\'ed\'eral de Lausanne, Switzerland\\
\idiapref Idiap Research Institute, Martigny, Switzerland\\
{\tt\small angel.martinez@idiap.ch, michael.villamizar@idiap.ch, odobez@idiap.ch}
}

\maketitle
\thispagestyle{empty}

\begin{abstract}
We propose to leverage Transformer architectures for non-autoregressive human motion prediction.
Our approach
decodes elements in
parallel from a query sequence, instead of conditioning on previous 
predictions such as in \soa RNN-based approaches.
%
In such a way our approach is less computational intensive and 
potentially avoids error accumulation to long term elements
in the sequence.
In that context, our contributions are fourfold:
(i) we frame human motion prediction as a sequence-to-sequence problem and
propose a non-autoregressive Transformer to infer the sequences of poses in parallel;
%
(ii) we propose to decode sequences of 3D poses from a query sequence
generated in advance with elements from the input sequence;
(iii) we propose to perform skeleton-based activity classification from the
encoder memory, in the hope that identifying the activity can improve predictions;
(iv) we show that despite its simplicity, our approach achieves competitive results in two public datasets,
although surprisingly more for short term predictions rather than for long term ones. 
\end{abstract}

\vspace*{-3mm}

\section{Introduction}
\pdfoutput=1 
\label{sec:introduction}
An important ability of an artificial system aiming at human behaviour understanding resides
in its capacity to apprehend the human motion,
including the possibility to anticipate motion and activities 
(\eg reaching towards objects).
As such, human motion prediction finds  applications in visual surveillance or human-robot
interaction (HRI) and  has been a hot topic researched for decades.

With the recent popularity of deep learning, Recurrent Neural Networks (RNN) have
replaced conventional methods that relied on Markovian 
dynamics~\cite{Lehrmann_CVPR_2014}
and smooth body motion~\cite{Sigal:IJCV:11} and instead learn these sequence 
properties from data.
However, motion prediction remains a challenging task due to the non-linear 
nature of the articulated body structure.
Although the different motions of the body landmarks are highly correlated, 
their relations and temporal evolution are hard to model in learning systems.
%

\begin{figure}[t]
\centering
\includegraphics[width=0.4\textwidth]{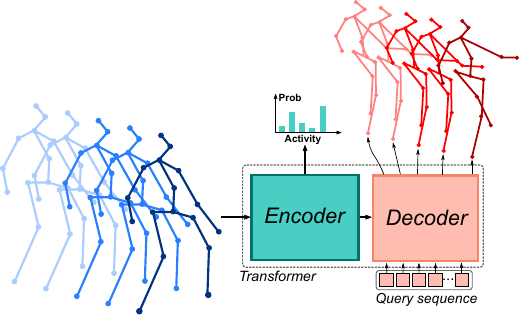}
\vspace{-0.3cm}
\caption{
	Proposed approach for non-autoregressive motion prediction approach with
	Transformers.
	The input sequence is encoded with the Transformer encoder.
	The decoder works in a non-autoregressive fashion generating the 
	predictions of all poses in parallel.
	Finally, the encoder embeddings are used for skeleton-based activity 
	classification.
      }
      \vspace{-0.3cm}
\label{fig:potr}
\end{figure}

Recently, a family of RNN-based approaches have proposed to frame the task of 
human motion prediction as a sequence-to-sequence problem.
These methods usually rely on
stacks of LSTM or GRU modules and solve the task 
with autoregressive decoding: generating predictions one at a time
conditioned on previous
predictions~\cite{julieta2017motion, Aksan_2019_ICCV}.
%
This practice has two major shortcomings.
First, autoregressive models are prone to accumulate prediction errors over time:
%
predicted elements are conditioned to previous predictions, containing a 
degree of error, thus potentially increasing the error for new predictions.
Second, autoregressive modelling is not parallelizable
which may cause deep models to be more computationally intensive
since predicted elements are generated sequentially one at time.

Since their breakthrough in machine translation~\cite{Vaswani_NIPS_2017}, Transformer neural network
have  been adopted in other research areas for different sequence-to-sequence tasks such as 
automatic speech recognition~\cite{katharopoulos_et_al_2020} and object 
detection~\cite{Carion_ECCV_2020}.
These methods leverage the long range memory of the attention modules to identify
specific entries in the input sequence which are relevant for prediction, 
a shortcoming of RNN models.
During training, Transformers allows parallelization 
with \emph{look ahead} masking.
Yet, at testing time, they use an autoregressive setting which makes it difficult
to leverage the parallelization capabilities.
Hence, autoregressive Transformers exhibit large inference processing times
hampering their use in applications that require real-time performance such
as in HRI.

In this paper, we thus investigate the use of non-autoregressive human motion prediction aiming to reduce 
computational cost of autoregressive inference with the Transformer neural
network and potentially avoid error propagation.
Our work is inline with recent methods~\cite{Carion_ECCV_2020, Jiatao_ICLR_2018} 
that perform non-autoregressive (parallel) decoding with Transformers.
Contrary to \soa methods that rely only in a Transformer encoder for human motion 
prediction~\cite{Aksan_2021, wei2020his}, our approach uses as well
a Transformer decoder architecture with self- and encoder-decoder 
attention.
Inspired by recent research in non-autoregressive machine 
translation~\cite{Jiatao_ICLR_2018}, we generate the inputs to the decoder 
in advance with elements from the input sequence.
We show that this strategy, though simple, is effective and
helps reducing the error in short and long term horizons.

In addition, we explore the inclusion of activity information by predicting 
as well activity from the input sequences.
Modelling motion and activity prediction jointly  has not often been 
investigated by previous works, though these topics are highly related.
Indeed, a better ability at identifying an activity may improve the selection
of the dynamics to be applied to the sequence. 
Hence, we propose a skeleton-based activity classification by 
classifying activities using the encoder self-attention predictions.
%
We train our models jointly for activity classification and motion prediction
and study the potential of this multi-task framework.
Code and models will be made public~\footnote{\url{https://github.com/idiap/potr}}.

The rest of the paper is organized as follows.
Section~\ref{sec:state-of-the-art} presents relevant \soa methods
to our work.
Section~\ref{sec:potr-method} introduces our approach for non-autoregressive 
motion prediction with Transformers.
Experimental protocol and results are presented in Section~\ref{sec:potr-results} and 
Section~\ref{sec:conclusions} concludes our work.
%
%

\pdfoutput=1 
\section{Related work}
\label{sec:state-of-the-art}
%
%
%
%
%

\mypartitle{Deep autoregressive methods.}
Early deep learning approaches used stacks of RNN units to model 
human motion.
For example, the work in~\cite{Fragkiadaki_ICCV_2015} introduces an 
encoder-recurrent-decoder (ERD) network with a stack of LSTM units.
%
%
The approach prevents of error accumulation and catastrophic drift by including
a schedule to add Gaussian noise to the inputs and increase the model robustness.
However, this scheduling is hard to tune in practice.
The work presented in~\cite{julieta2017motion} uses a encoder-decoder
RNN architecture with a single GRU unit.
The architecture includes a residual connection between
decoder inputs and outputs as a way of modeling velocities in the
predicted sequence.
This connectivity reduces discontinuity between input sequences and
predictions and adds robustness at long time horizons.
Along this line, the approach in~\cite{Aksan_2019_ICCV} introduce a 
decoder that explicitly models the spatial dependencies between the 
different body parts
 with small
specialized networks, each predicting a specific body part (\eg elbow).
Final predictions are decoded following the hierarchy of the body 
skeleton which reduces the drift effect.
Recently, a family of methods prevent the drift issue by including 
adversarial losses
and enhance prediction quality  with geodesic body
measurements~\cite{Gui_2018_ECCV} or by framing motion prediction as an
pose inpainting process with GANs~\cite{Hernandez_ICCV_2019}.
However, training with adversarial loses is difficult and hard to
stabilize.

%
%
%
%

Attention-based approaches have recently gained interest for modeling 
human motion.
For example, the work presented in~\cite{wei2020his} exploits a 
self-attention module to attend the input sequence with a sliding 
window of small subsequences from the input.
Ideally, attention should be larger in elements of the input sequence that
repeat with time.
Prediction works in an autoregressive fashion using a Graph Convolutional
Network (GCN) to decode attention embeddings to 3D skeletons.
Along the same line~\cite{Aksan_2021} introduces an spatio-temporal 
self-attention module to explicitly model the spatial components of the 
sequence.
Input sequences are processed by combining two separate self-attention modules: 
a spatial module to model body part relationships and a temporal module to
model temporal relationships.
Predictions are generated by aggregating attention embeddings with feedforward
networks in an autoregressive fashion.

Our work differs from these works.
First, our architecture is a encoder-decoder Transformer, with self-
and encoder-decoder attention.
This allows us to exploit the Transformer decoder to 
identify elements in the input sequence relevant for prediction.
Secondly, our architecture works in non-autoregressive fashion to prevent
the overhead of autoregressive decoding.

\mypartitle{Non-autoregressive modelling.}
Most neural network-based models for sequence-to-sequence modelling
use autoregressive decoding: generating entries in the sequence one at a time
conditioned on previous predicted elements.
This approach is not parallelizable causing deep learning models to be more
computationally intensive, as in the case of machine 
translation with Transformers~\cite{Vaswani_NIPS_2017,radford2019language}.
Although in principle Transformers are paralellizable, autoregressive decoding makes impossible
to leverage this property during inference.
Therefore, recent efforts have sought to parallelize decoding with transformers
in machine translation using \emph{fertilities}~\cite{Jiatao_ICLR_2018} and in visual object 
detection by decoding sets~\cite{Carion_ECCV_2020}.

Non-autoregressive modeling has also been explored in the human motion
prediction literature.
Clearly, the most challenging aspect is to represent the temporal dependencies
for decoding predictions.
Most of the solutions in the literature provide additional information to the
decoder that account for the temporal correlations in the target sequence.
Different methods have been proposed relying in decoder architectures that exploit 
temporal convolutions~\cite{Li_2018_CVPR}, feeding the decoder with learnable
embeddings~\cite{Li_TIP_2021}, or relying in a representation of the
sequence in the frequency domain~\cite{wei2019motion}.
The work presented in~\cite{wei2019motion} represents the 
temporal dependencies using the Discrete Cosine Transform (DCT) of the sequence.
During inference a GCN predicts the DCT coefficients of the target sequence.
However, to account for smoothness, during training, the GCN is trained to predict
both input and target sequence DCT coefficients.
The approach in~\cite{Li_2018_CVPR} performs a similar approach, 
modelling separately short term and long term dependencies with temporal convolutions.
%
Their decoder is composed of a short term and long term temporal encoders that move in 
a sliding window.
Short and long term information are then processed by a spatial decoder to produce
pose frames.

Our approach contrast from these methods in different ways.
First, we do not incorporate any prior information of the temporality of the
sequences and let the Transformer learn these from sequences of skeletons.
Additionally, our decoding process relies in a simple strategy to
generate query sequences from the inputs rather than
relying in learnable query embeddings.

\pdfoutput=1 
\section{Method}
\label{sec:potr-method}
\begin{figure*}
\centering
\includegraphics[width=0.9\linewidth]{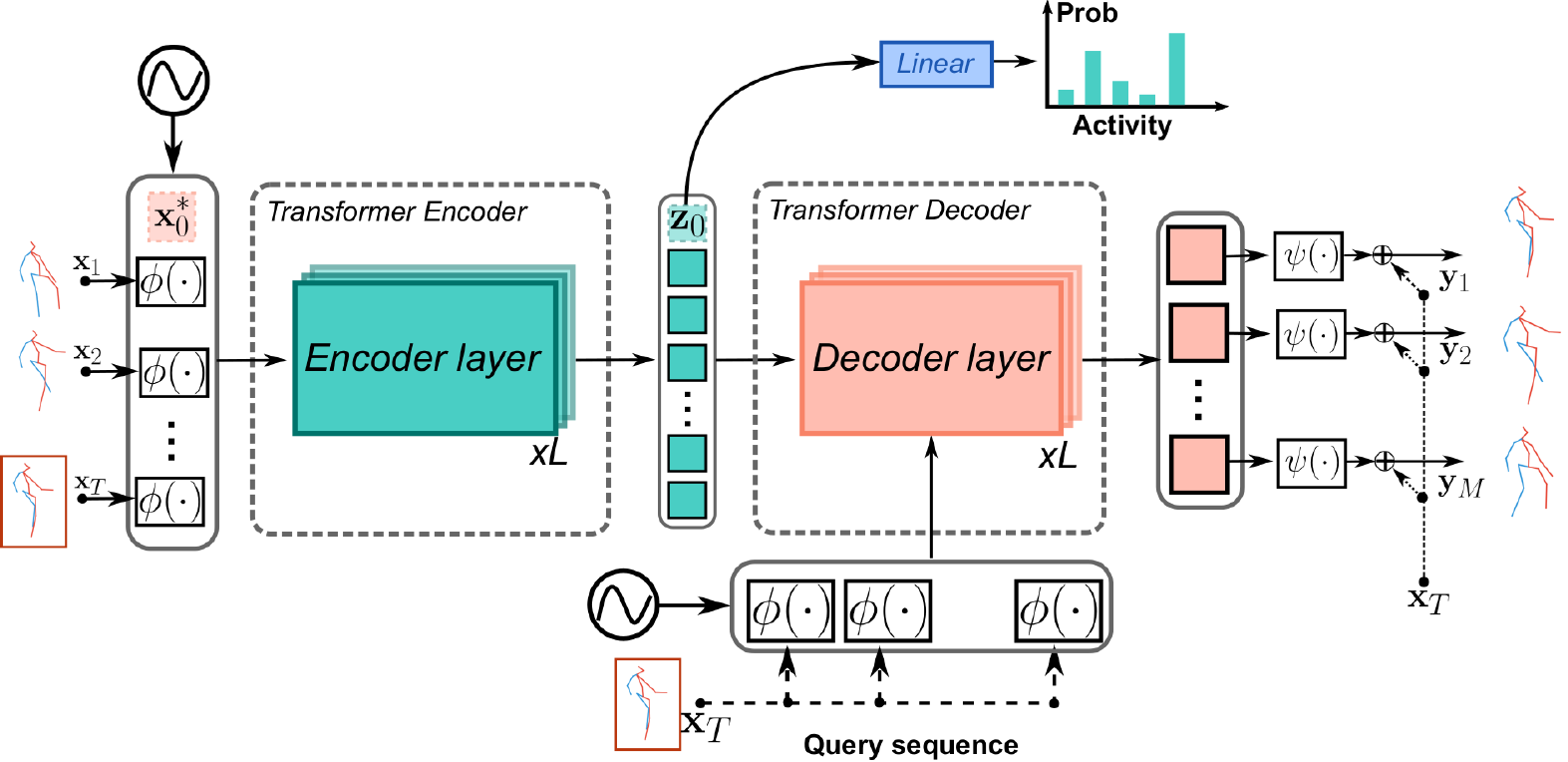}
\vspace{-0.3cm}
\caption{
	Overview of our approach for non-autoregressive human motion prediction.
	Our model is composed of networks $\PoseEmbFn$ and $\PoseDecFn$, and
	a non-autoregressive Transformer built on feed forward networks and 
 	multi-head attention layers as in~\cite{Vaswani_NIPS_2017}.
 	First, a network $\PoseEmbFn$ computes embeddings for each pose in the input 
 	sequence.
 	%
 	Then, the Transformer processes the sequence and decodes attention embeddings
 	in parallel.
 	%
 	%
 	Finally, the predicted sequence is generated with network $\PoseDecFn$ in a 
 	residual fashion.
 	%
 	Activity classification is performed by adding a learnable \emph{class token}
 	$\mathbf{x}_0$ to the input sequence.
}
\label{fig:potr}
\end{figure*}
%
%
The goal of our study is to explore solutions for human motion prediction
%
leveraging the parallelism properties of Transformers
during inference.
%
%
%
%
In the following sections we introduce our Pose Transformer (\potr), a 
non-autoregressive Transformer for motion prediction and skeleton-based activity 
recognition.

\subsection{Problem Formulation}
Given a sequence
$\mathbf X=\{\InputPose_{1:\InputLength}\}$ of 3D poses, we seek to predict the 
most likely immediate following
sequence $\mathbf Y=\{\PredictedPose_{1:\TargetLength}\}$, 
where $\InputPose_t, \PredictedPose_t\in\mathbb{R}^{\PoseDimension}$ are 
$N$-dimensional pose vectors (skeletons).
This problem is strongly related with conditional sequence modelling where 
the goal is to model the probabilities
$P(\mathbf Y | \mathbf X ; \TransParams)$ with model parameters $\TransParams$.
In our work, $\TransParams$ are the parameters of a Transformer.

%
Given its temporal nature, motion prediction has been widely addressed as an
autoregressive approach in an encoder-decoder configuration:
the encoder takes the conditioning motion sequence $\InputPose_{1:\InputLength}$
and computes a representation $\SelfAttnEmb_{1:\InputLength}$ (memory).
The decoder then generates pose vectors $\PredictedPose_t$ one by one taking 
$\SelfAttnEmb_{1:\InputLength}$ and its previous generated vectors 
$\PredictedPose_{\tau<t}$.
While this autoregressive approach explicitly models the temporal dependencies of
the predicted sequence $\PredictedPose_{1:\TargetLength}$, it requires to execute 
the decoder $\TargetLength$ times.
This becomes computationally expensive for very large Transformers, which 
in principle have the property of parallelization (exploited during training).
Moreover, autoregressive modelling is prone to propagate errors to future predictions:
predicting pose vector $\PredictedPose_t$ relies in predictions $\mathbf y_{\tau<t}$
which in practice contain a degree of error.
We address these limitations by modelling the problem in a non-autoregressive
fashion as we describe in the following.
%

\subsection{Pose Transformers}
\label{sec:potr-architecture}
%
%
The overall architecture of our \potr approach is shown in Figure~\ref{fig:potr}.
Similarly to the original Transformer~\cite{Vaswani_NIPS_2017}, our encoder and 
decoder modules are composed of feed forward networks and multi-head attention
modules.
While the encoder architecture stays unchanged,
the decoder works in a non-autoregressive fashion to avoid error
accumulation and reduce computational cost.

Our \potr comprises three main components: a pose encoding neural network 
$\PoseEmbFn$ that
computes pose embeddings for each 3D pose vector in the input sequence,
a non-autoregressive Transformer, and a pose decoding
neural network $\PoseDecFn$ that computes
a sequence of 3D pose vectors.
While the Transformer learns the temporal dependencies,
the networks $\PoseEmbFn$ and $\PoseDecFn$ shall identify spatial
dependencies between the different body parts for encoding and decoding
pose vector sequences.

More specifically, our architecture works as follows.
First, the pose encoding network $\PoseEmbFn$ computes an embedding of dimension 
$D$ for each pose vector in the input sequence $\InputPose_{1:\InputLength}$.
The Transformer encoder takes the sequence of pose embeddings (agreggated with
positional embeddings) and computes the representation
$\SelfAttnEmb_{1:\InputLength}$ with a stack of $\NumLayers$ multi-head 
self-attention layers.
The Transformer decoder takes the encoder outputs $\SelfAttnEmb_{1:\InputLength}$
as well as a \emph{query sequence}
$\QueryPose_{1:\TargetLength}$ and computes an output embedding with a 
stack of $\NumLayers$ multi-head self- and encoder-decoder attention layers.
Finally, pose predictions are generated in parallel by the network $\PoseDecFn$
from the decoder outputs and a residual connection with $\QueryPose_{1:\TargetLength}$.
%
%
%
%
We detail each component in the following.
%

%
%

\mypartitle{Transformer Encoder.}
It is composed of $\NumLayers$ layers, 
each with a standard architecture consisting of multi-head
self-attention modules and a feed forward networks.
%
The encoder receives as input the sequence of pose embeddings of dimension
$\EmbDimension$ added with positional encodings
and produces a sequence of embeddings $\SelfAttnEmb_{1:\InputLength}$ of the same
dimensionality.

\mypartitle{Transformer Decoder.}
Our Transformer decoder follows the standard architecture:
it comprise $\NumLayers$ layers of multi-head self- and encoder-decoder 
attention modules and feed forward networks.
%
In our work, every layer in the decoder generates predictions.
%
The decoder receives a query sequence $\QueryPose_{1:\TargetLength}$ and encoder
outputs $\SelfAttnEmb_{1:T}$ and produces $\TargetLength$ output embeddings in a
single pass.
These are then decoded by the network $\PoseDecFn$ into 3D body skeletons.

The decoding process starts by generating the input to the decoder 
$\QueryPose_{1:\TargetLength}$.
As remarked in~\cite{Jiatao_ICLR_2018} given that non-autoregressive decoding 
exhibits complete conditional independence between predicted elements 
$\PredictedPose_t$,
the decoder inputs should account as much as possible for the time correlations
between them.
Additionally, $\QueryPose_{1:\TargetLength}$ should be easily inferred.
%
%
Inspired by non-autoregressive machine translation~\cite{Jiatao_ICLR_2018},
we use a simple approach filling $\QueryPose_{1:\TargetLength}$
using copied entries from the encoder inputs.
More precisely, each entry $\QueryPose_t$ is a copy of a selected 
\emph{query pose} from the encoder inputs $\InputPose_{1:\InputLength}$.
We select the last element of the sequence $\InputPose_\InputLength$ as 
the query pose and fill the query sequence with this entry.
%
Given the residual learning setting, predicting motion can be seen as predicting
the necessary pose offsets from last conditioning pose $\InputPose_\InputLength$
to each element $\PredictedPose_t$.
%
%

\subsection{Pose Encoding and Decoding}
\label{sec:potr-enc-dec}
Input and output sequences are processed from and to 3D pose vectors with
networks $\PoseEmbFn$ and $\PoseDecFn$ respectively.
The network $\PoseEmbFn$ is shared by the Transformer encoder and decoder.
It computes a representation of dimension $\EmbDimension$ for each of the 3D 
skeletons in the input and query sequences.
The decoding network $\PoseDecFn$ transforms the $\TargetLength$ decoder predictions 
of dimension $\EmbDimension$ to 3D skeletons residuals independently at every 
decoder layer.
%
%

The aim of the $\PoseEmbFn$ and $\PoseDecFn$ networks is to model the spatial 
relationships between the different elements of the body structure.
To do this, we investigated two approaches.
In the first one we consider a simple approach setting $\PoseEmbFn$ 
and $\PoseDecFn$ with single linear layers.
%
%
In the second approach we follow~\cite{wei2020his} and 
use Graph Convolutional Networks (GCN) that 
densely learn the spatial connectivity in the body.

To make our manuscript self contained, we briefly introduce how GCNs 
work in our human motion prediction approach.
Given a feature representation of the human body with $\NumNodes$ nodes,
a GCN learns the relationships between nodes with
the strength of the graph edges represented by the adjacency matrix
$\AdjMatrix \in \mathbb{R}^{\NumNodes\times \NumNodes}$.
Examples of representations are body skeletons or embeddings.
A GCN layer $l$ takes as input a matrix of node features 
$\NodeFeats_{l-1}\in\mathbb{R}^{\NumNodes\times \NumNodeFeats}$
with $\NumNodeFeats$ features per node, and a set of learnable weights 
$\NodeWeights_l\in\mathbb{R}^{\NumNodeFeats\times \NodeNumFeatsOut}$.
Then, the layer computes output features

\begin{equation}
	\NodeFeats_{l} = \sigma(\AdjMatrix_l\mathbf \NodeFeats_{l-1}\NodeWeights_l),
\label{eq:gcn-layer}
\end{equation}

where $\sigma$ is an activation function.
A network is composed by stacking layers which aggregates features
of the vicinity of the nodes.

\begin{figure}[t]
\centering
\includegraphics[width=0.8\linewidth]{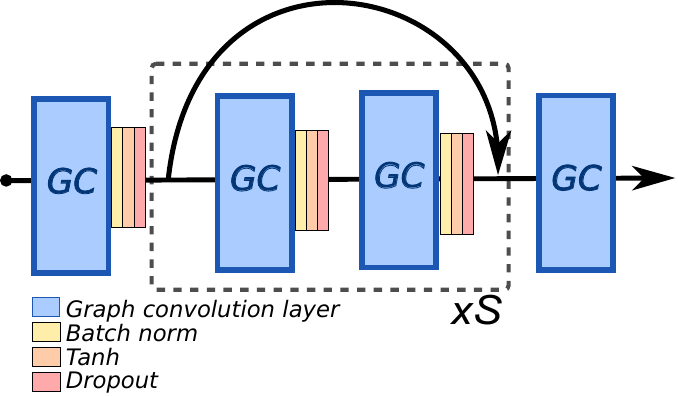}
\vspace{-0.2cm}
\caption{
	Our Graph Convolutional Network architecture.
	It comprises graph convolution layers followed by \emph{tanh}
	activations, batch normalization, and dropout layers.
	As in~\cite{wei2020his}, our architecture has $S$ 
	residual connections.
	%
}
\label{fig:gcn}
\end{figure}

Our GCN architecture is shown in Figure~\ref{fig:gcn}.
It is inspired in the architecture presented in~\cite{wei2020his}, where
matrices $\AdjMatrix_l$ and weights $\NodeWeights_l$ are learnt.
It is composed of a stack of $S$ residual modules of
graph convolution layers followed by batch normalization,
\emph{tanh} activations and dropout layers.
%
%
%
%
We set the internal feature dimension to  $\NumNodeFeats=512$ features per node 
until the output layer that generates pose embeddings.
%
Though we can normally squeeze as many inner layers, we set $S=1$.

\subsection{Activity Recognition}
\label{sec:potr-activity}
Activity can normally be understood as a sequence of motion of the different
body parts in interaction with the scene context (objects or people).
In our method, the Transformer encoder encodes the body motion with a series
of self-attention layers.
%
%
We explore the use of encoder outputs $\SelfAttnEmb_{1:\InputLength}$ for 
activity classification (as a second task)
and train a classifier to determine
the action corresponding to the motion sequence presented as input to the
Transformer.
%

%
%
We explore two approaches.
The first approach consist on using the entire Transformer 
encoder outputs $\SelfAttnEmb_{1:\InputLength}$ as input to the classifier.
However, these normally contain many zeroed entries suppressed
by the probability maps normalization in the multi-head attention layers.
Naively using these for activity classification might lead our classifier to 
struggle in discarding these many zero elements.
Therefore, similar to~\cite{dosovitskiy2020}, we include a 
specialized \emph{class token} in the input sequence to store information about
the activity of the sequence.
The class token $\InputPose_0$ is a learnable embedding that is padded to 
input sequence to form $\InputPose_{0:\InputLength}$.
In the output of encoder embeddings $\SelfAttnEmb_{0:\InputLength}$,
$\SelfAttnEmb_0$ works as the activity representation of the encoded motion sequence.
To perform activity classification we feed $\SelfAttnEmb_0$ to a single linear layer
to predict class probabilities for $C$ activity classes
(see Figure~\ref{fig:potr}).

\subsection{Training}
We train our model in a multi-task fashion to jointly predict motion and
activity.
Let $\hat{\PredictedPose}_{1:\TargetLength}^l$ be the predicted sequence of 
$\PoseDimension$-dimensional
pose vectors at layer $l$ of the Transformer decoder.
We compute the layerwise loss

\begin{equation}
 L_{l} = \frac{1}{\TargetLength\cdot \PoseDimension} \sum_{t=1}^{\TargetLength} ||\hat{\PredictedPose}_{t}^l - \PredictedPose_{t}^{*}||_{1},
\label{eq:pose-loss}
\end{equation}

where $\PredictedPose_{t}^{*}$ is the ground truth skeleton at target sequence entry $t$.
The overall motion prediction loss $L_{motion}$ is computed by averaging the losses
over all decoder layers $L_l$.
%
%
Finally, we train our \potr with the loss

\begin{equation}
L_{POTR} = L_{motion} + \lambda L_{activity},
\label{eq:potr-loss}
\end{equation}

where $L_{activity}$ is the multi-class cross entropy loss and $\lambda=1$.

\pdfoutput=1 
\section{Experiments}
\label{sec:potr-results}
%
%
This section presents the experiments we conducted to evaluate our approach.

\subsection{Data}
\mypartitle{Human 3.6M~\cite{H36M}.}
We used the Human 3.6M dataset in our experiments for human motion 
prediction.
The dataset depicts seven actors performing 15 activities, \eg walking, eating, 
sitting, \etc.
We followed standard protocols for training and 
testing~\cite{julieta2017motion, Aksan_2019_ICCV, wei2019motion}.
Subject 5 is used for testing while the others for training.
Input sequences are 2 seconds long and testing is performed over the first
400 ms of the predicted sequence.
%
Evaluation is done in a
total of 120 sequences (8 seeds) across all activities by
computing the Euler angle error between predictions and ground truth.
%
%

\mypartitle{NTU Action Dataset~\cite{Shahroudy_2016_NTURGBD}.}
The NTU-RGB+D dataset is one of the biggest 
benchmark datasets for human activity recognition.
It is composed of 58K Kinect~2 videos of 40 different actors performing 60
different actions.
%
We followed the cross subject evaluation protocol provided by the authors
using 40K sequences for training and 16.5K for testing.
%
%
Given the small length of the sequences, we set input and output sequences length
to 1.3 seconds~(40 frames) and 660 ms~(20 frames) respectively.
\subsection{Implementation details}
\mypartitle{Data Preprocessing}.
We apply standard normalization to the input and ground truth skeletons by
substracting the mean and dividing by the standard deviation computed over the whole training set.
For the H3.6M dataset we remove global translation of the skeletons
and represent the skeletons with rotation matrices.
%
Skeletons in the NTU dataset are represented in 3D coordinates and are centred
by subtracting the spine joint.

\mypartitle{Training}.
We use Pytorch as our deep learning framework in all our experiments.
Our \potr is trained with AdamW~\cite{loshchilov2018decoupled} setting
the learning rate to $10^{-04}$ and weight decay to $10^{-05}$.
\potr models for the H3.6M dataset are trained during 100K steps with warmup schedule
during 10K steps.
For the NTU dataset we train \potr models during 300K steps with warmup schedule
during 30K.
%

\mypartitle{Models}.
We set the dimension of the embeddings in our \potr models to  $\EmbDimension=128$.
The multi-head attention modules are set with pre-normalization
and four attention heads and four layers in encoder and decoder.
%
%

\subsection{Evaluation metrics}
\mypartitle{Euler Angle Error}.
We followed standard practices to measure the error of pose predictions in the H3.6M 
dataset
by computing the euclidean norm between predictions and ground truth
Euler angle representations.
%
%
%

\mypartitle{Mean Average Precision (mAP)}.
We use mAP@10cm to measure the performance of predictions in the NTU dataset.
%
A successful detection is considered when the predicted 3D body landmark falls within
a distance less than 10 cm from the ground truth.
%
%
%

\mypartitle{Mean Per Joint Position Error  (MPJPE)}.
%
%
We use the MPJPE to evaluate error in the NTU dataset.
MPJPE measures the average error in Euclidean distance between the predicted 3D body
landmarks and the ground truth.
%
%

\subsection{Results}

\subsubsection{Evaluation on H3.6M Dataset}
%
%
In this section, we validate our proposed approach for motion prediction 
in the H3.6M dataset.
%

\mypartitle{Non-Autoregressive Prediction}.
Table~\ref{tab:hm36-arvsnonar} compares the performance in terms of the Euler
angle error of our \potr with its autoregressive version (\potr-AR).
Lower values are better.
%
The autoregressive version do not use the query pose and predicts pose vectors one 
at a time from its own predictions.
%
%
Our non-autoregressive approach shows lower error than its counter part
in most of the time intervals.
%
%

\begin{table}
\centering
\resizebox{\linewidth}{!}{\begin{tabular}{l | c  c  c  c  c  c }

milliseconds & 80 & 160 & 320 & 400 & 560 & 1000 \\

\hline

\potr-AR            & 0.23 & 0.57 & 0.99 & 1.14 & 1.37 & 1.81 \\

\potr 		  & 0.23 & \textbf{0.55} & \textbf{0.94} & 1.08 & 1.32 & 1.79 \\

\Xhline{2\arrayrulewidth}

\potr-GCN (enc) &\textbf{0.22} & 0.56 & \textbf{0.94} & \textbf{1.01} & \textbf{1.30} & \textbf{1.77} \\

\potr-GCN (dec) & 0.24 & 0.57 & 0.96 & 1.10 & 1.33 & 1.77 \\

\potr-GCN (full) & 0.23 & 0.57 & 0.96 & 1.10 & 1.33 & 1.80 \\

\hline

\end{tabular}
}
\caption{\textbf{H3.6M} prediction performance in terms of the Euler
		angle error.
		Top: autoregressive (\potr-AR) and non-autoregressive \potr models
		using linear layers for networks $\PoseEmbFn$ and $\PoseDecFn$.
		Bottom: non-autoregressive models with GCNs for network
		$\PoseEmbFn$ (enc), network  $\PoseDecFn$ (dec) and both (full).
		%
		%
}
\label{tab:hm36-arvsnonar}
\end{table}

\mypartitle{Pose Encoding and Decoding}.
We experimented with the networks $\PoseEmbFn$ and $\PoseDecFn$
using either linear layers or GCNs.
Table~\ref{tab:hm36-arvsnonar} reports the results (bottom part).
We indicate when models are trained with GCN in the
encoder (enc), decoder (dec) or in both (full).
%
%
%
%
We observe that the use of GCN reduces the errors when it is applied exclusively to 
the encoder.
Using a shallow GCN ($S=1$) $\PoseDecFn$ might be a weak attempt to decode pose vectors.
However, we observed that the small size of the H3.6M dataset might not be
enough to learn deeper GCN architectures.
%

\begin{figure*}
\centering
\includegraphics[width=\textwidth]{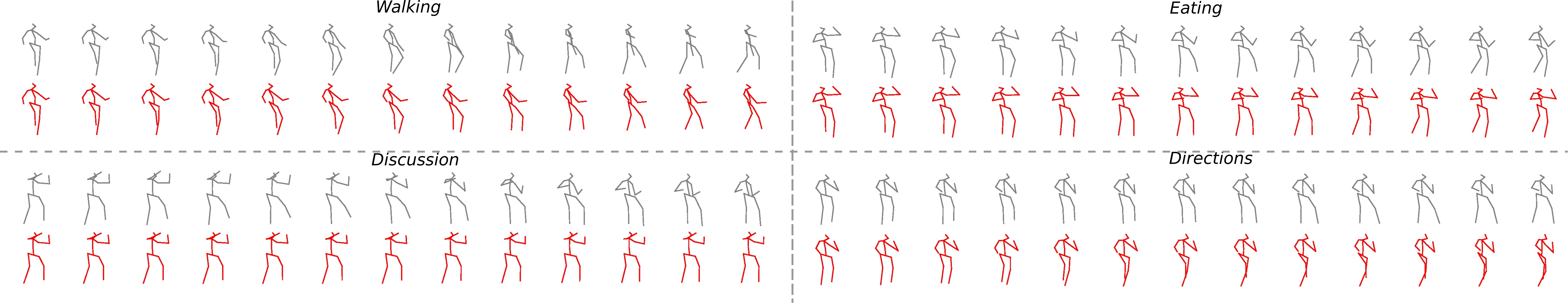}
\vspace{-0.7cm}
\caption[Qualitative results for the H36M dataset.]{
	Qualitative results for the \textbf{H36M} dataset.
	We show results for four actions and show ground truth and predicted elements
	coloured in gray and red respectively.
	%
}
\label{fig:h36m-visualization}
\end{figure*}

\begin{table*}[!h]
\centering
\resizebox{\textwidth}{!}{\begin{tabular}{l | c c c c | c c c c | c c c c | c c c c }

 &  \multicolumn{4}{c|}{Walking} & \multicolumn{4}{c|}{Eating} & \multicolumn{4}{c|}{Smoking} & \multicolumn{4}{c}{Discussion}\\

\hline

milliseconds & 80 & 160 & 320 & 400  & 80 & 160 & 320 & 400 & 80 & 160 & 320 & 400 & 80 & 160 & 320 & 400 \\

\hline

Zero Velocity~\cite{julieta2017motion}   & 0.39 & 0.68 & 0.99 & 1.15  & 0.27 & 0.48 & 0.73 & 0.86 & 0.26 & 0.48 & 0.97 & 0.95 & 0.31 & 0.67 & 0.94 & 1.04 \\

Seq2seq.~\cite{julieta2017motion}      & 0.28 & 0.49 & 0.72 & 0.81 &
0.23 & 0.39 & 0.62 & 0.76 &
0.33 & 0.61 & 1.05 & 1.15 &
0.31 & 0.68 & 1.01 & 1.09\\

AGED~\cite{Gui_2018_ECCV} & 0.22 & \underline{0.36} & 0.55 & 0.67 &
0.17 & \textbf{0.28} & 0.51 & 0.64 &
0.27 & 0.43 & \textbf{0.82} & 0.84 &
0.27 & 0.56 & \textbf{0.76} & \textbf{0.83}\\

RNN-SPL~\cite{Aksan_2019_ICCV} & 0.26 & 0.40 & 0.67 & 0.78 &
0.21 & 0.34 & 0.55 & 0.69 &
0.26 & 0.48 & 0.96 & 0.94 &
0.30 & 0.66 & 0.95 & 1.05 \\

DCT-GCN (ST)~\cite{wei2020his} & \underline{0.18} & \textbf{0.31}& \textbf{0.49} & \textbf{0.56} &
\underline{0.16}& \underline{0.29} & \underline{0.50} & \underline{0.62} &
\underline{0.22} & \underline{0.41} & 0.86 & \textbf{0.80} &
0.20 & \textbf{0.51} & \underline{0.77} & \underline{0.85}\\


ST-Transformer~\cite{Aksan_2021} & 0.21 & \underline{0.36} & \underline{0.58} & \underline{0.63} & 
0.17 & 0.30 & \textbf{0.49} & \textbf{0.60} &
\underline{0.22} & 0.43 & 0.88 & \underline{0.82} & 
\underline{0.19} & \underline{0.52} & 0.79 & 0.88 \\

\Xhline{2.5\arrayrulewidth}
\potr-GCN (enc)  & \textbf{0.16} & 0.40 & 0.62 & 0.73 &
\textbf{0.11} & \underline{0.29} & 0.53 & 0.68 &
\textbf{0.14} & \textbf{0.39} & \underline{0.84} & \underline{0.82} &
\textbf{0.17} & 0.56 & 0.85 & 0.96 \\

\Xhline{2\arrayrulewidth}

\end{tabular}}
\caption{
	\textbf{H3.6M} performance comparison with the \soa in terms of the
	Euler angle error for the common
	\emph{walking}, \emph{eating}, \emph{smoking} and \emph{discussion}
	across different horizons.
	%
}
\label{tab:hm36-soa1}
\end{table*}
\mypartitle{Comparison with the State-Of-The-Art.}
Tables~\ref{tab:hm36-soa1} and~\ref{tab:hm36-soa2} compares our best
performing model with the 
\soa in terms of angle error for all the activities in the dataset.
%
%
Our \potr often obtains the first and second lower errors in in the short term, 
and the lowest average error in the 80ms range.
The use of the last input sequence entry as the query pose most probably helps 
to significantly  reduce the error in the immediate horizons.
However, this strategy introduces larger errors for longer horizons where the
difference between further pose vectors in the sequence and the query pose is larger
(see Figure~\ref{fig:h36m-visualization} for some examples).
In such a case, it appears that autoregressive approaches perform better as a result of updating
the conditioning decoding distribution to elements closer in time.
%
%

\begin{figure*}
\centering
	(a)\includegraphics[width=0.9\textwidth]{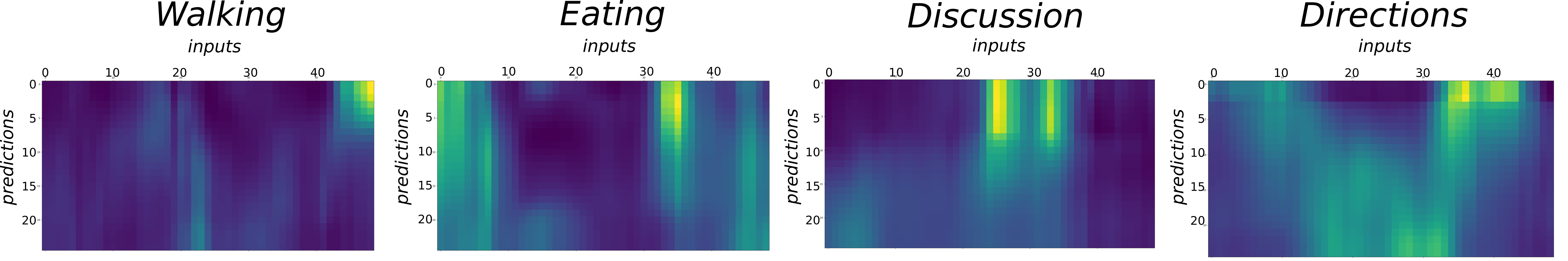}\\
	(b)\includegraphics[width=0.9\textwidth]{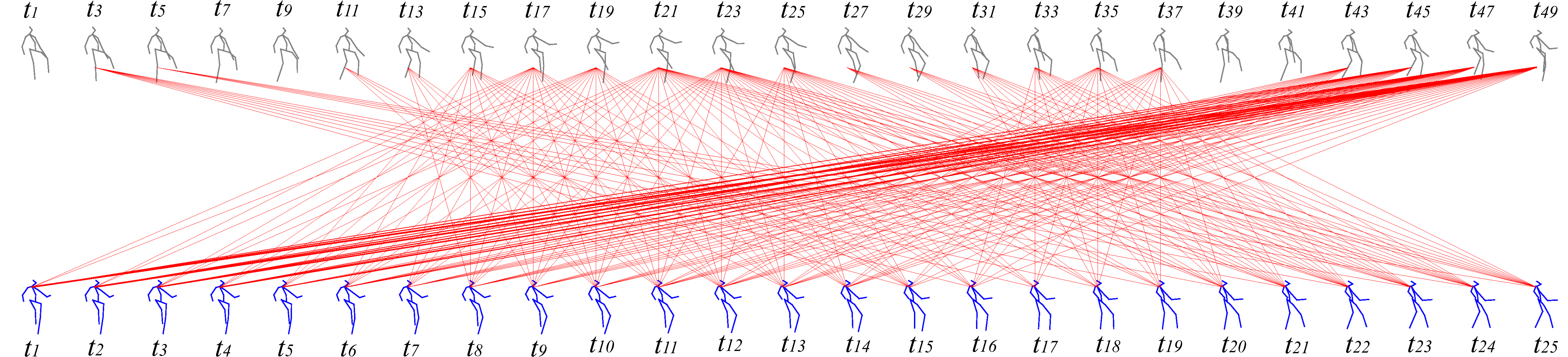}
\vspace{-0.2cm}
\caption[\textbf{H3.6M} dataset encoder-decoder attention weight visualization.]{
	 \textbf{H3.6M} datasest encoder-decoder attention weight visualization.
	 (a) Raw encoder-decoder attention maps. Input and predicted entries are 
	 represented by columns and rows respectively.
 	 (b) Attention weights between input (gray) and predicted (blue) skeleton 
 	 sequences of the \emph{walking} action. Only weights larger than the median 
 	 are visualized.
	 The thickness of the lines are proportional to the attention weights.
	 For visualization purposes we show only half of the input sequence;

}
\label{fig:h36m-attn-visualization}
\end{figure*}

\mypartitle{Attention Weights Visualization.}
In Figure~\ref{fig:h36m-attn-visualization}(a) we visualize the 
encoder-decoder attention maps for one predicition instance of four activities in the dataset.
Figure~\ref{fig:h36m-attn-visualization}(b) further shows the attention between 
elements of the input and predicted sequences for the \emph{walking} action.
%
%
%
%
Due to the continuity within such activity, we can notice a high dependency (attention)
between the last elements of the input and the firts elements of the predicted sequences,
while the prediction of further elements also pay attention to other input elements
of the input matching the same phase of the walking cycle.
A similar behavior is observed for the direction example. 
For the eating and discussion activities involving less body motion,
we can notice that while the approach slightly attends to the last elements
of the input, it also strongly attends to other specific segments.
Further analysis would be needed to analyse the behavior of these weight matrices. 

%
%

\mypartitle{Computational Requirements.}
We measured the computational requirements of models \potr and \potr-AR by the
number of sequences per second (SPS) of their forward pass in a single Nvidia
card GTX 1050.
We tested models with 4 layers in encoder and decoder, and 4 heads in their
attention layers.
We input sequences of 50 elements and predict sequences of 25 elements.
\potr runs at 149.2 SPS while
\potr-AR runs at 8.9 SPS.
Therefore, the non-autoregressive approach is less computationally intensive.

\begin{table*}[t]
\centering
\resizebox{\textwidth}{!}{\begin{tabular}{l | >{\centering}p{0.45cm}>{\centering}p{0.45cm}>{\centering}p{0.45cm}>{\centering}p{0.45cm} | >{\centering}p{0.45cm}>{\centering}p{0.45cm}>{\centering}p{0.45cm}>{\centering}p{0.45cm} | >{\centering}p{0.45cm}>{\centering}p{0.45cm}>{\centering}p{0.45cm}>{\centering}p{0.45cm} | >{\centering}p{0.45cm}>{\centering}p{0.45cm}>{\centering}p{0.45cm}>{\centering}p{0.45cm} | >{\centering}p{0.45cm}>{\centering}p{0.45cm}>{\centering}p{0.45cm}>{\centering}p{0.45cm} | >{\centering}p{0.45cm}>{\centering}p{0.45cm}>{\centering}p{0.45cm}>{\centering\arraybackslash}p{0.45cm}}

 &  \multicolumn{4}{c|}{Directions} & \multicolumn{4}{c|}{Greeting} & \multicolumn{4}{c|}{Phoning} & \multicolumn{4}{c}{Posing} & \multicolumn{4}{c|}{Purchases} & \multicolumn{4}{c}{Sitting} \\

\hline

milliseconds & 80 & 160 & 320 & 400  & 80 & 160 & 320 & 400 & 
80 & 160 & 320 & 400 & 80 & 160 & 320 & 400 & 
80 & 160 & 320 & 400  & 80 & 160 & 320 & 400 \\

\hline

Seq2seq~\cite{julieta2017motion} & 
0.26 & 0.47 & 0.72 & 0.84 & 
0.75 & 1.17 & 1.74 & 1.83 &  
\underline{0.23} & \underline{0.43} & \underline{0.69} & \underline{0.82} &  
0.36 & 0.71 & 1.22 & 1.48& 
0.51 & 0.97 & 1.07 & 1.16 & 
0.41 & 1.05 & 1.49 & 1.63 \\

AGED~\cite{Gui_2018_ECCV} & 
\underline{0.23} & \underline{0.39} & \textbf{0.63} & \textbf{0.69} & 
0.56 & 0.81 & 1.30 & 1.46 & 
\textbf{0.19}& \textbf{0.34} & \textbf{0.50} & \textbf{0.68} & 
0.31 & 0.58 & 1.12 & 1.34 & 
0.46 & 0.78 & \textbf{1.01} & \textbf{1.07} & 
0.41 & 0.76 & 1.05 & 1.19 \\

DCT-GCN (ST)~\cite{wei2020his}  & 
0.26 & 0.45 & \underline{0.71} & \underline{0.79} & 
0.36 & \textbf{0.60} & \textbf{0.95} & \textbf{1.13} & 
0.53 & 1.02 & 1.35 & 1.48 & 
\underline{0.19} & \textbf{0.44} & \textbf{1.01} & \textbf{1.24} & 
\underline{0.43} & \underline{0.65} & 1.05 & 1.13 & 
\underline{0.29} & \textbf{0.45} & \textbf{0.80} & \textbf{0.97}\\


ST-Transformer~\cite{Aksan_2021}  & 
0.25 & \textbf{0.38} & 0.75 & 0.86 & 
\underline{0.35} & \underline{0.61} & \underline{1.10} & 1.32 & 
0.53 & 1.04 & 1.41 & 1.54 & 
0.61 & 0.68 & \underline{1.05} & \underline{1.28} & 
\underline{0.43} & 0.77 & 1.30 & 1.37 & 
\underline{0.29} & \underline{0.46} & \underline{0.84} & \underline{1.01} \\

\Xhline{2\arrayrulewidth}

\potr-GCN (enc) & 
\textbf{0.20} & 0.45 & 0.79 & 0.91 & 
\textbf{0.29} & 0.69 & 1.17 & \underline{1.30} & 
0.50 & 1.10 & 1.50 & 1.65 & 
\textbf{0.18} & \underline{0.52} & 1.18 & 1.47 &  
\textbf{0.33} & \textbf{0.63} & \underline{1.04} & \underline{1.09} & 
\textbf{0.25} & 0.47 & 0.92 & 1.09 \\

\Xhline{2\arrayrulewidth}

\end{tabular}}

\vspace{0.3cm}

\resizebox{\textwidth}{!}{\begin{tabular}{l | >{\centering}p{0.45cm}>{\centering}p{0.45cm}>{\centering}p{0.45cm}>{\centering}p{0.45cm} | >{\centering}p{0.45cm}>{\centering}p{0.45cm}>{\centering}p{0.45cm}>{\centering}p{0.45cm} | >{\centering}p{0.45cm}>{\centering}p{0.45cm}>{\centering}p{0.45cm}>{\centering}p{0.45cm} | >{\centering}p{0.45cm}>{\centering}p{0.45cm}>{\centering}p{0.45cm}>{\centering}p{0.45cm} | >{\centering}p{0.45cm}>{\centering}p{0.45cm}>{\centering}p{0.45cm}>{\centering}p{0.45cm} | >{\centering}p{0.45cm}>{\centering}p{0.45cm}>{\centering}p{0.45cm}>{\centering\arraybackslash}p{0.45cm}}

 &   \multicolumn{4}{c|}{Sitting down} & \multicolumn{4}{c}{Taking photos} &  \multicolumn{4}{c|}{Waiting} & \multicolumn{4}{c|}{Walking Dog} & \multicolumn{4}{c|}{Walking Together} & \multicolumn{4}{c}{Average}\\

\hline

milliseconds & 80 & 160 & 320 & 400 & 
80 & 160 & 320 & 400 & 80 & 160 & 320 & 400  & 
80 & 160 & 320 & 400 & 80 & 160 & 320 & 400 & 
80 & 160 & 320 & 400\\

\hline

Seq2seq.~\cite{julieta2017motion} & 0.39 & 0.81 & 1.40 & 1.62 & 
0.24 & 0.51 & 0.90 & 1.05 & 
0.28 & 0.53 & 1.02 & 1.14 & 
0.56 & 0.91 & 1.26 & 1.40 & 
0.31 & 0.58 & 0.87 & 0.91 & 
0.36 & 0.67 & 1.02 & 1.15\\

AGED~\cite{Gui_2018_ECCV} & 
0.33 & \underline{0.62} & \underline{0.98} & \underline{1.10} & 
0.23 & 0.48 & 0.81 & 0.95 & 
0.24 & \textbf{0.50} & 1.02 & \textbf{1.13} & 
0.50 & 0.81 & 1.15 & \textbf{1.27} & 
0.23 & 0.41 & \underline{0.56} & \underline{0.62} & 
0.31 & \underline{0.54} & \underline{0.85} & \underline{0.97}\\

DCT-GCN (ST)~\cite{wei2020his}  & 
\underline{0.30} & \textbf{0.61} & \textbf{0.90} & \textbf{1.00} & 
\underline{0.14} & \textbf{0.34} & \textbf{0.58} & \textbf{0.70} & 
0.23 & \textbf{0.50} & \textbf{0.91} & \underline{1.14} & 
0.46 & \underline{0.79} & \textbf{1.12} & \underline{1.29} & 
\textbf{0.15} & \textbf{0.34} & \textbf{0.52} & \textbf{0.57} & 
\underline{0.27} & \textbf{0.52} & \textbf{0.83} & \textbf{0.95}\\ 


ST-Transformer~\cite{Aksan_2021} & 
0.32 & 0.66 & \underline{0.98} & \underline{1.10} & 
\underline{0.15} &  \underline{0.38} & \underline{0.64} & \underline{0.75} & 
\underline{0.22} & \underline{0.51} & \underline{0.98} & 1.22 & 
\underline{0.43} & \textbf{0.78} & \underline{1.15} & 1.30 & 
\underline{0.17} & \underline{0.37} & 0.58 & \underline{0.62} & 
0.30 & 0.55 & 0.90 & 1.02\\

\Xhline{2\arrayrulewidth}

\potr-GCN (enc) & \textbf{0.25} & 
0.63 & 1.00 & 1.12 & \textbf{0.12} & 
0.41 & 0.71 & 0.86 & 
\textbf{0.17} & 0.56 & 1.14 & 1.37 & 
\textbf{0.35} & \underline{0.79} & 1.21 & 1.33 & 
\textbf{0.15} & 0.44 & 0.63 & 0.70 & 
\textbf{0.22} & 0.56 & 0.94 & 1.01\\ 

\Xhline{2\arrayrulewidth}

\end{tabular}}

\caption{
	Euler angle error results for the reminder of the 11 actions in the \textbf{H3.6M} dataset
	with our main non-autoregressive transformer.
}
\label{tab:hm36-soa2}
\end{table*}
%

\subsubsection{Evaluation on NTU Dataset}
This section presents our results on jointly predicting motion and activity 
on the NTU dataset.

\begin{table}
\centering
\resizebox{\linewidth}{!}{\begin{tabular}{l | c  c  c  c c c | c | c }

milliseconds & 80 & 160 & 320 & 400 & 500 & 660 & avg & accuracy\\

\hline

\potr-AR            & 0.96 & 0.92 & 0.85 & 0.83 & 0.80 & 0.76 & 0.76 & 0.32\\

 \potr		  	   & 0.96 & \textbf{0.93} & \textbf{0.89} & \textbf{0.87} & \textbf{0.86} & \textbf{0.84} & \textbf{0.84} & \textbf{0.38} \\

\potr ($\lambda=0$)	  & 0.96 & \textbf{0.93} & \textbf{0.89} & \textbf{0.87} & 0.85 & 0.83 & 0.83 & - \\ 

\potr (memory)	  & 0.96 & 0.92 & 0.88 & \textbf{0.87} & 0.85 & 0.83 & 0.83 & 0.30\\

\Xhline{2\arrayrulewidth}

\potr-GCN (enc) & 0.96 & 0.92 & 0.88 & \textbf{0.87} & 0.85 & 0.83 & 0.83 & 0.27\\

\potr-GCN (dec) & 0.96 & 0.92 & 0.88 & 0.86 & 0.85 & 0.83 & 0.83 & 0.34\\

\potr-GCN (full) & 0.95 & 0.90 & 0.85 & 0.84 & 0.82 & 0.79 & 0.79 & 0.30\\

\hline

\end{tabular}
}	

\caption{
\textbf{NTU} motion prediction performance in terms of the mAP@10cm
for different time horizons.
%
Higher values are better.
Model marked with \emph{memory} replace the class token with the encoded memory 
for activity classification.
%
%
}
\label{tab:ntu-motion}
\end{table}
\begin{figure}[!h]
\centering
\includegraphics[width=0.45\linewidth]{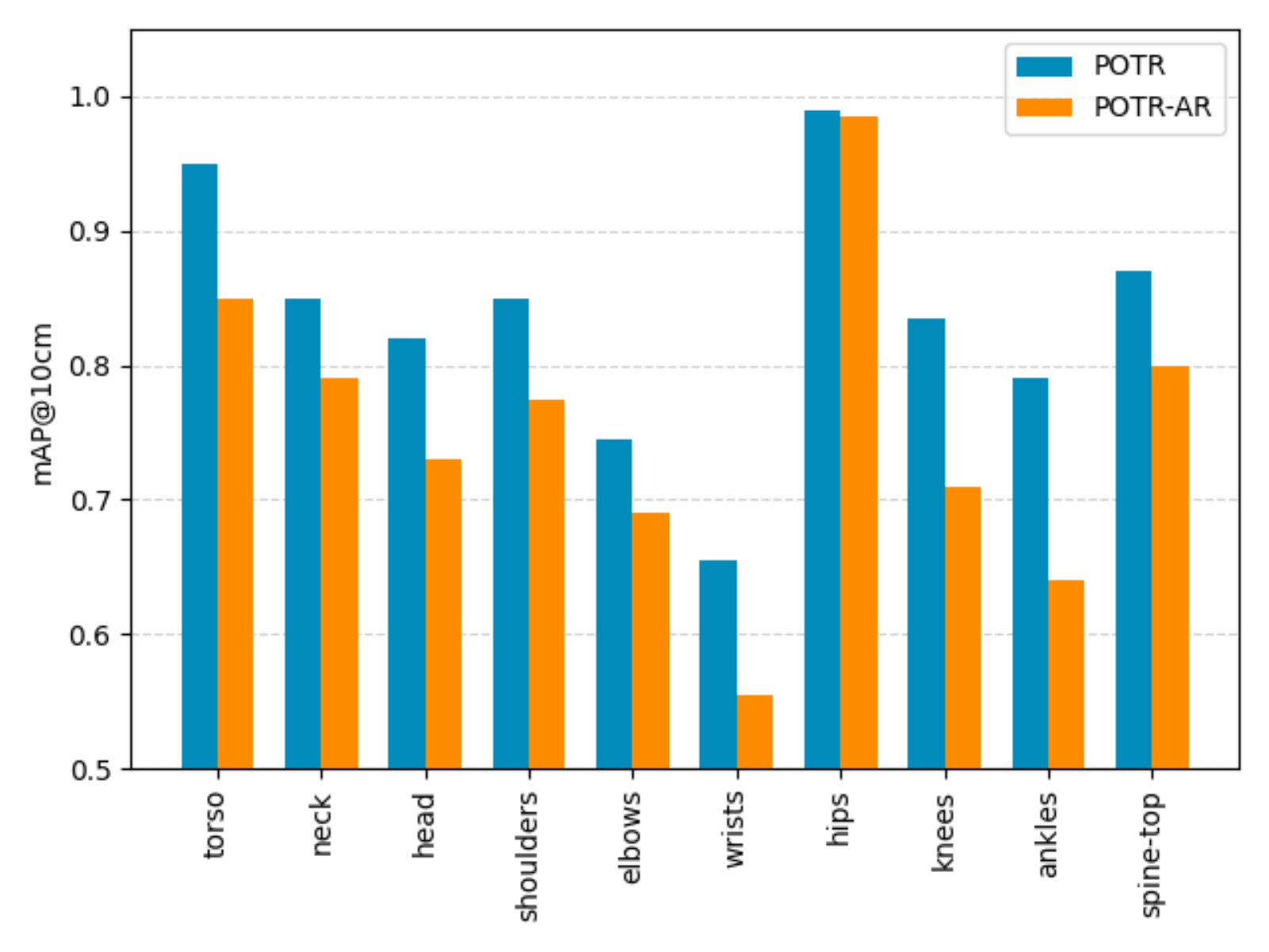}\ 
\includegraphics[width=0.45\linewidth]{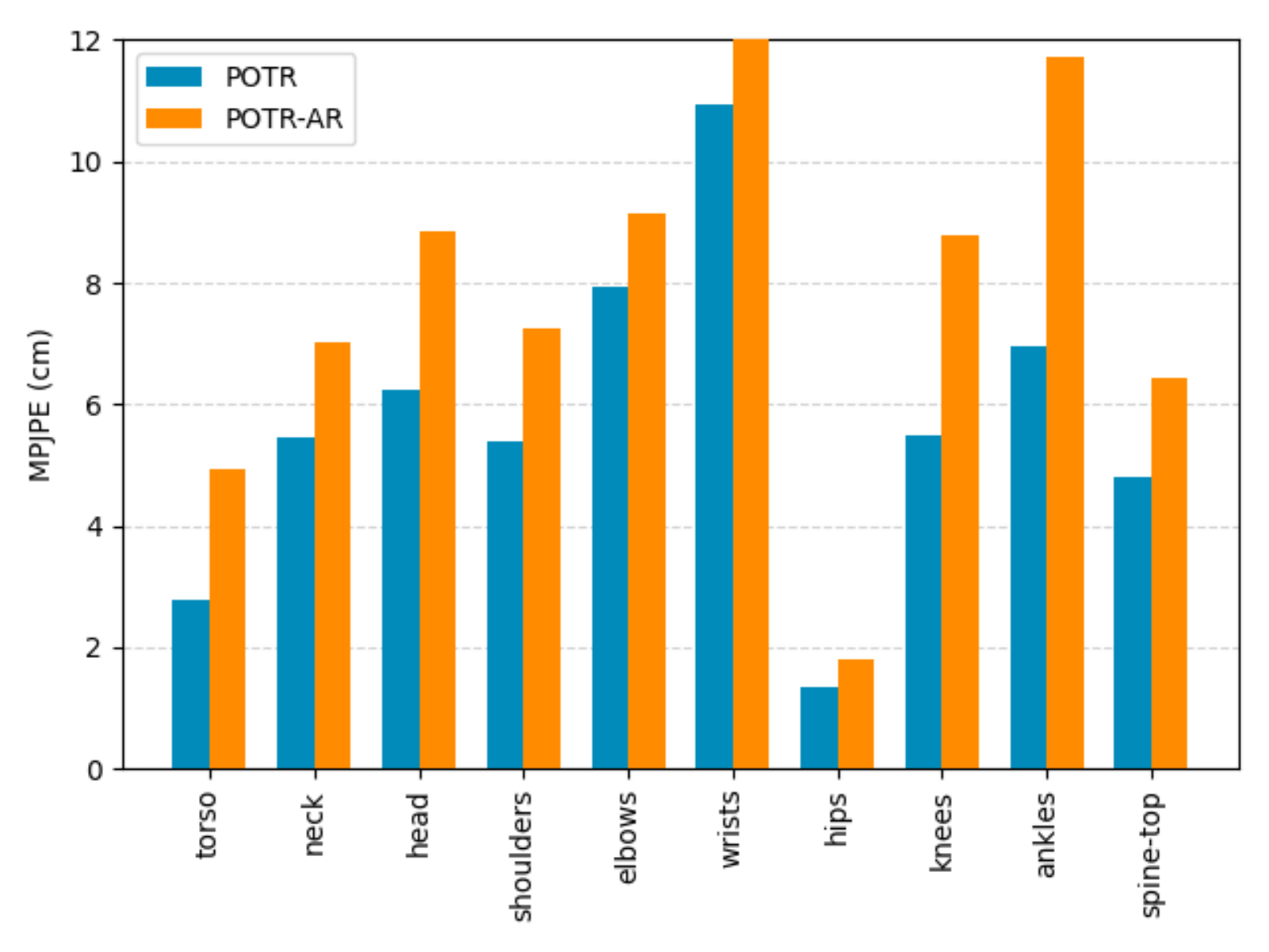}
\vspace{-0.3cm}
\caption{
	\textbf{NTU} per body part motion prediction performance in terms of (a) mAP@10cm.
	Higher is better and, (b) MPJPE. Lower is better.
}
\label{fig:ntu-per-joint}
\end{figure}
%
\mypartitle{Motion Prediction Performance}.
Table~\ref{tab:ntu-motion} compares our \potr with the different decoding 
settings using the mAP.
%
Notice that removing the activity loss ($\lambda=0$) slightly drops the
performance for the longer horizons.
%
The non-autoregressive setting shows higher mAP than the autoregressive
setting, specially in long term.
%
However, setting the networks $\PoseEmbFn$ and $\PoseDecFn$ with GCNs does not 
bring many benefits compared to using linear layers.
%

Figure~\ref{fig:ntu-per-joint} compares their per body part mAP and MPJPE
using linear layers for $\PoseEmbFn$ and $\PoseDecFn$.
%
%
\potr-AR shows larger MPJPE and lower 
mAP than \potr specially for the body extremities (arms and legs).

\mypartitle{Activity Recognition}.
%
%
Table~\ref{tab:ntu-motion}
compares the classification accuracy for the different \potr configurations.
Using a specialized activity token shows better performance than using
the encoder memory $\SelfAttnEmb_{1:\InputLength}$.
Given that the memory embeddings contain many 
non-informative zeroed values the classifier could get stuck in an attempt to 
ignore them.
%
%
%
%
%

Table~\ref{tab:ntu-action} compares the classification accuracy with \soa methods
from sequences of 3D skeletons or color images.
We can see that our approach only performs inline with the \soa 
method with the lowest accuracy, 
but can note that methods using only skeletal information perform worse. 
Among this category, the method presented by~\cite{Shahroudy_2016_NTURGBD}
achieves the largest accuracy.
It relies on a stack of LSTM modules with specialized part-based cells 
that processes groups of body parts (arms, torso and legs).
Such an explicit scheme could potentially improve our approach which
is a simpler modeling of the overall body motion,
especially given the size of the training set.
%
The best performance overall is obtained by~\cite{Diogo_CVPR_2018} which combines
color images and skeleton modalities.
In such case, including image context provides extra information
that cannot be extracted from skeletal data, \eg
objects of interaction.

\begin{table}
\centering
\begin{tabular}{l | c  c  | c }

Method & Skeletons & RGB & Accuracy\\

\hline

Skeletal quads~\cite{Georgios_ICPR_2014} & $\surd$ & - & 38.62 \% \\		

2 Layer P-LSTM~\cite{Shahroudy_2016_NTURGBD} & $\surd$ & - & 62.93 \% \\

Multi-task~\cite{Diogo_CVPR_2018} & $\surd$ & $\surd$ & \textbf{85.5} \% \\

Multi-task~\cite{Diogo_CVPR_2018} & - & $\surd$ & 84.6 \% \\

\Xhline{2\arrayrulewidth}

Ours \potr & $\surd$ & - & 38.0 \% \\

Ours \potr (memory) & $\surd$ & - & 30.0 \%

\end{tabular}
\vspace*{-2mm}
\caption{
Activity classification performance comparison with the \soa in the \textbf{NTU}
dataset.
We specify if methods work with skeleton sequences or color images.
}
\vspace*{-2mm}
\label{tab:ntu-action}
\end{table}
%

\section{Conclusions}
\label{sec:conclusions}
In this paper we addressed the problem of non-autoregressive human motion
prediction with Transformers.
We proposed to decode predictions in parallel from a query sequence generated 
in advance with elements from the input sequence.
Additionally, we analyzed different settings to encode and decode pose 
embeddings.
We leveraged the encoder memory embeddings to perform activity 
classification with an activity token.
Our non-autogressive method outperforms its autoregressive version in long
term horizons and is less computationally intensive.
Finally, despite the simplicity of our approach we have obtained competitive results 
on motion prediction in two public datasets.

Our work opens the door for more research.
One of the main drawbacks in our method is the increased error at long term 
horizons as a consequence of non-autoregressive decoding and relying in
a single pose vector as query sequence.
A more suitable strategy to explore would be  to rely in a set of $M$ query poses
by sampling from the input or selected
using the encoder self-attention embeddings by position modelling such as
in~\cite{Jiatao_ICLR_2018}.
%
%
%
%

\mypartitle{Acknowledgments}:
This work was supported by the European Union under the EU Horizon 2020 Research
and Innovation Action MuMMER (MultiModal Mall Entertainment Robot), project ID
688146, as well as the Mexican National Council for Science and Technology~(CONACYT)
under the PhD scholarships~program.

{\small
\bibliographystyle{ieee_fullname}
\bibliography{biblio}
}

\end{document}